\pdfoutput=1
\documentclass[11pt]{article}

\usepackage[final]{coling}

\usepackage{times}
\usepackage{latexsym}

\usepackage[T1]{fontenc}
\usepackage[utf8]{inputenc}

\usepackage{microtype}

\usepackage{inconsolata}

\usepackage{graphicx}

\title{Annotating Compositionality Scores for Irish Noun Compounds is Hard Work}

\author{
  \textbf{Abigail Walsh\textsuperscript{1}},
  \textbf{Teresa Clifford\textsuperscript{1}},
  \textbf{Emma Daly\textsuperscript{1}},
  \textbf{Jane Dunne\textsuperscript{1}}, \and
  \textbf{ \large Brian Davis\textsuperscript{1}},
  \textbf{ \large Gearóid Ó Cleircín\textsuperscript{2}}
  \\
  \textsuperscript{1}ADAPT Centre, Dublin City University, 
  \textsuperscript{2}Fiontar \& Scoil na Gaeilge, Dublin City University \\
    \small{
    \textsuperscript{1}\textbf{firstname.lastname@adaptcentre.ie} %\href{mailto:email@domain}{email@domain}
    \textsuperscript{2}\textbf{firstname.lastname@dcu.ie}
    }
}

\begin{document}
\maketitle
\begin{abstract}
Noun compounds constitute a challenging construction for NLP applications, given their variability in idiomaticity and interpretation. In this paper, we present an analysis of compound nouns identified in Irish text of varied domains by expert annotators, focusing on compositionality as a key feature, but also domain specificity, as well as familiarity and confidence of the annotator giving the ratings. Our findings and the discussion that ensued contributes towards a greater understanding of how these constructions appear in Irish language, and how they might be treated separately from English noun compounds.
\end{abstract}

\section{Introduction}

Red tape, black market, gold standard: noun compounds (NCs) such as these pose challenges for many Natural Language Processing (NLP) tasks, including machine translation, information retrieval, text summarization, and others. The shades of meaning encapsulated in the above multi word expressions demonstrate a challenge to the principle of compositionality,\footnote{"the meaning of a (syntactically complex) whole is a function only of the meanings of its (syntactic) parts together with the manner in which these parts were combined"} and make noun compounds a compelling linguistic feature to analyse in an NLP context. 
%Like many linguistic concepts, this may best be explained through examples. 
%For example, the NC \textit{dath bán}, compromising of the word \textit{dath} `colour' and \textit{bán} `white' is considered fully compositional. Both words are functioning as expected and they explain the meaing of the construction without the need of outside or real world knowledge. In contrast to this, the NC Mac tíre. It directly translates to son of the land but it means wolf. This is considered a fully non compositional NC; the meaning of the NC could not be derived from the two words and neither word is functioning as expected. Things that are to be considered when examining and attempting to annotate for compositionality include the real world knowledge required to understand a NC; a human will understand that a fire alarm is an alarm that goes off in the presence of fire as opposed to an alarm that produces fire because of real world understanding. One must also take into account that words can have multiple meanings or meanings that are ambiguous. This, combined with the fact that different annotators may have different understandings of words or NCs due to their familiarity, means that NCs may have different compositionality scores from different annotators.   

%NB: Add in the section numbers
In this paper, we examine the idiomaticity displayed by noun compounds (NCs) in Irish, building a corpus of varied domain text annotated with NCs and ratings of compositionality, domain specificity, annotator familiarity and confidence. A brief analysis is presented of the annotator judgements of these constructions, which are still largely unexplored for Irish. The annotation guidelines are presented, along with discussion on challenging cases which required particular attention. Annotation is still underway for the dataset, however the pilot task annotations are made available for public use, along with a collection of Irish noun compounds of various levels of compositionality (the first of its kind).

%[Brian]
%Motivation for considering different types of data
%(long-tail of linguistic features)
%Linguistic analysis of the language
%vs Computational approach

%(low resource languages, scarcity of language/linguistic features, evaluation of language technology)

\section{Background}
%Everything by Aline and also PARSEME. UniDive meeting is super important here. 

Understanding how complex linguistic expressions derive meaning from their constituent parts is crucial for the progress of modern Natural Language Processing (NLP) systems.  Noun compounds (NCs) provide an interesting avenue for investigation; testing the capabilities of models in capturing the different levels of compositionality inherent to NCs. This offers an opportunity to better understand how language is encoded in these models. Additionally, NCs display a relatively fixed syntactic expression, have a high-degree of variability, require knowledge-intensive interpretation, and are frequently found across languages \cite{calzolari-etal-2002-towards,girju-etal-2005-on}.

In \citet{cordeiro-etal-2019-unsupervised}, innovative methods for predicting the compositionality of compounds with unsupervised methods are introduced. Using distributional semantic models, they compare how well compositionality is captured in these vector embeddings when compared with human judgements, with the results showing idiomaticity correlates with human judgements for NCs in three languages.
%By utilizing word embeddings, the authors suggest an approach that taps into contextual information stored in vector representations. The focus on the aspect of their approach highlights its practicality and scalability when evaluating compositionality in linguistic datasets pushing the boundaries of unsupervised methods for comprehending and forecasting compositionality.
Following this, \citet{garcia-etal-2021-assessing} present a curated dataset of English and Portuguese NCs, designed to assess idiomaticity at both the type and token level. Their experiments compare the ability of BERT- \cite{vaswani-etal-2017-attention} and ELMo- \cite{peters-etal-2018-deep} based models representing NCs in varied contexts with idiomaticity ratings given by human annotators. The results indicate that vector representations struggle to capture the variability often displayed by  NCs. These experiments, along with others investigating the capacity of Large Language Models (LLMs) to handle high-level and abstract language features \cite{bisk-etal-2019-piqa,collins-etal-2022-structured,misra-etal-2023-comps} reveal areas where such models still struggle to attain human-level language performance.
%expressions within vector representations. This dataset, uniquely labeled at both type and token levels offers a nuanced understanding of the complexities associated with language. Through providing an evaluation framework the authors contribute to uncovering how idiomatic nuances are reflected in vector models. The dual-level labeling ensures a precise assessment of idiomatic elements within the vector space.

%These studies not only showcase advancements in approaches to evaluating compositionality but also emphasize the importance of specialized datasets. Additionally, they help us gain an insight into the challenges involved in working with nominal compound compositionality.

%PARSEME task (verbal MWEs) [Abigail]

For minority languages, such as Irish, assessing the capacity of LLMs is limited by the lack of data, linguistic analysis, and other essential resources \cite{lynn-2022-report}. However, with the increasing dominance of these models across NLP applications, such evaluations of model capacity in both monolingual and multilingual contexts will become increasingly important. We follow the example of \citet{reddy-etal-2011-empirical} and \citet{farahmand-etal-2015-multiword}, as well as those of the authors mentioned above, in developing a dataset of compositionality ratings for Irish NCs. Taking inspiration from the PARSEME Shared Tasks \cite{savary-etal-2017-parseme,ramisch-etal-2018-edition,ramisch-etal-2020-edition} we design annotation guidelines for the identification of noun compound candidates (NCCs) and annotate constructions in Irish data from varied domains, to better understand the distribution and features of NCs in the Irish language.
%With each new iteration, the scope broadens with the inclusion of different languages (27 languages in v1.2), and categories (8 full categories and 2 language-specific categories). The annotation guidelines (link) have been extended for each of these languages and categories, and issues are explored through pilot annotations and forum discussions. This format of developing annotation guidelines informed our process (described below).

%UniDive (less detail) [Abigail]
%The UniDive (Universality, Idiosyncrasy and Diversity in Language Technology) project is a COST-funded initiative for expanding the scope of language technology to address these areas often overlooked in favour of quickly improving tools and models for high-resource languages.  

\section{Defining NCs for Irish}

%The precise definition of noun compounds is difficult to determine, particularly in a multilingual context. (insert some definitions). 
Within the field of NLP, NCs are often treated in a similar fashion as other types of multiword expressions (MWEs) \cite{bauer-2019-compounds}, and there is some overlap with other constructions. For instance, \citet{baldwin-kim-2010-multiword} distinguish between NCs and nominal compounds, the latter including constructions with a non-nominal modifier (e.g. `connecting flight'). Both semantic and lexical idiosyncrasy are traits of NCs, but statistically idiosyncratic NCs may overlap with institutionalised phrases \cite{sag-etal-2002-multiword} (e.g. `traffic light'). Compound terminology (e.g. `hydrochloric acid') and named entities (e.g. `London Bridge') can appear similar to NCs; the former describe terms particular to some domain, and the latter are lexical units referring to a particular discrete entity \cite{nouvel-etal-2016-named}( usually a person, location or organisation). Definitions and terminology can vary, however, we consider a noun compound as any construction consisting of two or more words, where the head word is a noun, and the distribution of the construction is that of a noun.

%Challenges in defining noun compounds. Do they exist in Irish? Some examples clearly do, but they do not appear to be present as in English. Analyse some contrasting examples (mac tíre vs madra rua, fear ghrinn, terminology) Named entities.

Irish NCs were evaluated in the Universal Dependencies treebank by \citet{mcguinness-etal-2020-annotating}. Discussions around the proper annotation of these constructions highlighted the importance of non-compositionality as a distinguishing feature of NCs, with the absence of a definite article and the presence of a cranberry word being other indicating features. The first feature is generally applicable to all NC candidates, and so we focus our analysis of NCs through this lens.

%NB: maybe move this section
With our working definition of NCs, it remains an open question whether these constructions are a feature of the Irish language which is shared with English or other languages, or whether the syntactic structure of the Irish language lends itself to creating noun compound-like noun phrases (see Section \ref{sec:diff}).
%[NB: maybe rephrase this, would like to include a more in-depth analysis of noun phrase construction in Irish]
%The question of whether there is value in treating them differently in linguistic analysis or compiling lexica. One of the aims of this research was to gain perspective on the presentation of noun compounds in Irish, and their frequency in a variety of domains.

\subsection{Translation of English NCs}

As a complement to the annotation process, we analysed a list of two-word English NCs that had been translated into Irish. Out of 280 English NCs, only 30 NCs translated into equivalent two-word Irish NCs, with these being largely compositional compound terminology (e.g. \textit{cúntas banc} `bank account', \textit{ráta breithe} `birth rate'). Non-compositional NCs tended to translate directly into a single word (e.g. \textit{gealt} `nut case') or required a literal translation to capture the same meaning in Irish (e.g. \textit{umar na haimléise} (slough of-the wretchedness) `rock bottom').

\section{Annotation Process}

\subsection{Data and Preprocessing}
%[Teresa]
The corpora came from two different sources: 
The \textbf{Dúchas} dataset (Bailiúchán na Scol, Ábhar Co. na Gaillimhe) \cite{duchas} contains digital files from the Irish Folklore Collection; which are transcriptions of folkloric materials collected by primary school students between 1937 and 1939. The sentences are in dialectal Irish with non-standard orthography, and some resources are transcribed by the public, which can lead to spelling errors.\footnote{Creative Commons licence (CC BY-NC 4.0).}

The \textbf{Universal Dependencies Irish Dependency Treebank} (UD-IDT) \cite{lynn-foster-2016-universal} (v2.12) \cite{ud-idt} contains Irish in mixed domains (e.g. fiction, government, legal, news, web), originally sourced from \textit{Nua-Chorpas na hÉireann} and a corpus of Irish public administration translations. The treebank contains gold-standard dependency parsed trees and includes morphological features and part-of-speech (POS) information.\footnote{Creative Commons license (CC BY-SA 3.0).}

The INCEPTION platform \cite{klie-etal-2018-inception},\footnote{Available at \url{https://inception-project.github.io/}.} was used to carry out the annotation. The data was uploaded in CoNLL-U format, which is the format of the UD-IDT. The Dúchas dataset was automatically processed using a script to extract whole sentences from sentence blocks, which were parsed with UDPipe\footnote{Available at \url{https://lindat.mff.cuni.cz/services/udpipe/}} using the irish-idt-ud-2.12-230717 model.

\subsection{Training Annotators}
Recruiting and training annotators for this task presented challenges, in part because of the limited number of native speakers of Irish, most of whom are located in rural pockets of Irish-speaking or \textit{Gaeltacht} areas, and the lack of required linguistic expertise among these speakers (such challenges and the required planning to overcome them are described in the 
\href{https://assets.gov.ie/241755/e82c256a-6f47-4ddb-8ce6-ff81df208bb1.pdf}{Digital Plan for the
Irish Language Speech and Language Technologies 2023-2027}).

Three annotators were involved in the annotation process; two speakers at the C2 level, one of whom was a non-canonical Irish (L2) speaker (A1), and the other a Connemara Irish\footnote{The three major dialects of spoken Irish are Munster, Connemara or Connacht, and Donegal or Ulster, named for the regions that they are spoken in.} (L2) speaker comparable to a native speaker (A2). The third annotator had high language proficiency at the B2 level in Munster Irish (A3).

As the annotators had varying levels of language proficiency and linguistic training, a tailored approach was necessary, considering both the specific linguistic characteristics of the Irish language as well as a deep understanding of the problem. Three pilot tasks were designed to simultaneously develop the annotation guidelines and train annotators in scoring compositionality, domain specificity, and annotator familiarity consistently.

%Each noun compound (NC) was given a \textit{compositionality score} to indicate how transparent or opaque it is ranging from 0 (totally opaque), to 5 (totally transparent). Annotators also provided a \textit{confidence score} from 1 to 3 reflecting their certainty in their NC annotations.

%A \textit{domain specificity} score ranging from 1 to 3 indicated the level of domain specificity in the language, where 1 represents general domain language and 3 denotes very domain specific language. The \textit{annotator familiarity score} measured the annotators familiarity with the NC with a range of 1 (unfamiliar) to 3 (highly familiar).

%\subsection{IAA and disagreements}
%To assess annotator agreement several pilot tasks were conducted. 
%During pilot task 3, annotators were given both a 5-point and a 6-point scale to rate compositionality of NCs, allowing for comparison between scores of each NC on both scales. 
The level of agreement between annotators was calculated using Cohen's Weighted Kappa for each pair of annotators. The results of this analysis are summarized in Table \ref{tab:pt-1}, and provide a glimpse into the level of consensus among the annotators. %NB: check the annotators assigned to each score

\begin{table}[]
\centering
\begin{tabular}{l | r | r | r }
 &
  \multicolumn{1}{l}{\textbf{A1 v A2}} &
  \multicolumn{1}{l}{\textbf{A1 v A3}} &
  \multicolumn{1}{l}{\textbf{A2 v A3}} \\
Pilot 1                     & { 0.45} & { 0.64} & { 0.5}  \\
Pilot 2                     & { 0.54} & { 0.31} & { 0.3}  \\
Pilot 3 (6 pt)           & { 0.54} & { 0.42} & { 0.51} \\
Pilot 3 (5 pt)           & { 0.41} & { 0.42} & { 0.55} \\
Average                     & { 0.54} & { 0.49} & { 0.49}
\end{tabular}
\label{tab:pt-1}
\caption{Cohen's weighted kappa scores for three annotators across the three pilot tasks. A1 stands for Annotator 1, etc. Pilot 1 stands for Pilot Task 1, etc. Pilot 3 (6 pt) stands for Inter annotator agreement (IAA) of compositionality scores using the six-point scale in Pilot Task 3, while Pilot 3 (5 pt) compares IAA using the five-point scale. Average refers to the macro-averaged IAA score across Pilot 1, Pilot 2 and Pilot 3 (6 pt).}
\end{table}

\subsection{Annotation Guidelines}
Following the refinement process of the pilot tasks, the annotation guidelines were developed into a descriptive document for the task of annotating noun compound candidates (NCCs) with compositionality scores, domain-specificity scores, annotator-familiarity scores, and annotator confidence scores. Additionally, annotators marked when the NCC was also a named entity. The guidelines offer instructions pertaining to each of the following metrics/steps:

\paragraph{Determining the NCC:} The first step in annotation is to identify the noun compound candidate for additional annotation. It was decided early in the pilot tasks to limit our annotation of NCCs to constructions of only \textbf{two words}, these being contiguous words in a construction, one word of which was a noun or adjective dependent on the other word, a noun, and the construction must follow the distribution of a noun. We eliminated determiners from these constructions, and where a determiner was interleaved between the items of the construction, we did not consider this a valid candidate (e.g. \textit{mí na meala} (month of-the honey) `honeymoon') (see Section \ref{sec:diff}).%(see Example \ref{ex:det-1}).

%\begin{exe}
%    \ex \gll \textit{mí} \textit{\textbf{na}} \textit{meala} \\
%    month the honey[GEN] \\
%    \glt `honeymoon' 
%    \label{ex:det-1}
%\end{exe}

%This decision is discussed further in Section \ref{}.

\paragraph{Compositionality:} After selection of the NCC, a compositionality score is assigned from [0-5] (0 being totally non-compositional or opaque, 5 being totally compositional or transparent), scoring how closely each of the components of the construction behaves semantically within the construction when compared to either component occurring outside of this lexical context. For example, \textit{mac tíre} (son of-land) `wolf' was given a score of 0, \textit{mac léinn} (son of-learning) `student' was given a score of 1, and \textit{mac dearthár} (son of-brother) `nephew' would receive a score of 5.\footnote{The guidelines provide justification for the use of a six-point compositionality score, a decision which was explored during the pilot tasks.}

%\begin{exe}
%    \ex \gll \textit{mac} \textit{tíre} \\
%    son land[GEN] \\
%    `wolf'
%    \label{ex:comp-1}
%    \ex \gll \textit{mac} \textit{léinn} \\
%    son learning[GEN] \\
%    `student'
%    \label{ex:comp-2}
%    \ex \gll \textit{mac} \textit{dearthár} \\
%    son brother[GEN] \\
%    `nephew'
%    \label{ex:comp-3}
%\end{exe}

\paragraph{Domain Specificity:} During the pilot tasks, the significance of domain in determining the typical usage of a word became readily evident. Intuitively, a more domain-specific term will display less variability, i.e. the meaning will be more consistent, and will aid delineation between compound nouns and multiword terminology. A score between [1-3] is assigned, indicating the level of domain knowledge necessary to understand the NCC. For example, \textit{lucht éisteachta} `listener' requires no specialised knowledge to understand and, is so is scored 1, while \textit{lann ardluais} `high speed blade' is a term requiring domain expertise to understand, and is scored 3.

\paragraph{Annotator Familiarity:} Determining compositionality and domain specificity requires familiarity with the NCC. A score between [1-3] (unfamiliar to familiar) is therefore assigned based on how familiar annotators are with the NCC's meaning and usage. 

\paragraph{Confidence:} Difficult annotation cases can indicate a gap in the clarity of the guidelines, an unusual construction, a lack of information or knowledge, or some other factor. A confidence score of [1-3] is thus assigned to rate the confidence of the annotator in identifying or annotating the NCC.

\paragraph{Determining Named Entity:} The final step of annotation is to determine whether the NCC is also a named entity (NE) (i.e. whether it refers to a particular entity or not).

%\subsection{Description of dataset}
%data annotated
%publish the collection of noun compounds
%TBD on release of annotated dataset

%The dataset currently consists of a compilation of the pilot task data, revised according to the current annotation guidelines. Additionally, 200 sentences of the IUDT data have been annotated, with a total of [Y] NCCs identified, [Z] of which are NEs. In a collection of the annotated NCCs, [x1] of theses NCCs have a compositionality score between 0-2 (non-compositional), and [x2] of these NCCs have a compositionality score between 0-1 (very non-compositional). 

%31 of the annotated NCCs were added to the NC collection (needs a name). The NC collection consists of 165 entries, and will be shared publically.[footnote about the content being shared after the anonymous review is over] [we can publish this on gogs or similar]. The current version of the dataset will be shared publically (depending on the license again!).

\section{Preliminary Analysis} 
Annotation is still ongoing, however, some initial analysis and insights are presented on 200 sentences of the UD-IDT and 270 NCCs, with the pilot tasks containing a further 54 sentences across both datasets, and 105 NCCs. 

\subsection{Difficult Cases}
\label{sec:diff}
Certain elements of the Irish language and the data added to the difficulty of the annotation tasks. Some of these are explored below.
%While going through the first rounds of annotations and working towards the creation of annotation guidelines, we began to discover what elements of the Irish language and Irish language data made annotating NCCs for compositionality difficult. These ranged from difficulties in determining what was to be conisdered an NCC to extract, to difficulties caused by dialectal differences. The main difficult cases we ran into related to the defininte article, data standardisation, named enities and annotatpr familiarity.

\paragraph{Definite Articles:}
Irish language lacks an indefinite article \cite{christian-1999-graimear}, and definite nouns can be used for description (e.g. \textit{fear an phoist} (man of-the post) `post man') or possession (e.g. \textit{cat an fhir} (cat of-the man) the man's cat), with annotators agreeing the latter constructions should not be considered NCCs. However, some constructions were a challenge to this dichotomy; for example, \textit{toradh na talún} (fruit of-the land) `the fruits of the earth' could be either attributive (i.e. `earth fruits') or possessive (i.e. `the earth's fruits'). To avoid disagreements, constructions containing a definite article were not annotated.

\paragraph{Named Entities and Compositionality}

While NEs were also annotated as NCCs, difficulties arose when applying compositionality ratings to these constructions, particularly for place names. NEs such as \textit{Baile \textbf{Átha Cliath}} (town of-fords wattled) `\textbf{Dublin} City' were likely coined as descriptive of the area, and with the historical features that gave the area its name now missing, these names become non-compositional. Some place names, however, could be considered compositional in contemporary times (e.g. \textit{Béal Feirste} (mouth-of tidal-ford) `Belfast'), however the compositionality of such constructions is debatable, as the historical meaning is likely not intended to be applied by most language users. NEs were, as such, assigned compositionality scores of 0 by default, with the possibility of such constructions being removed in the future.

\paragraph{Annotator Familiarity:}
During the pilot tasks, differences in annotator knowledge led to disagreements in the annotation. Domain-specific NCCs such as \textit{measín aistriúchán} `machine translation' require domain knowledge to assess compositionality of the components and the construction. Additionally, due to variable language proficiency, annotators may be unfamiliar with less common constructions and misinterpret their meaning. For example, the construction \textit{i \textbf{riocht éan}} `in form-of bird' initially was interpreted by annotators as an NCC, however, the construction \textit{i riocht} actually forms the compound preposition `about to', and so \textit{riocht éan} should not be annotated as NCC. To analyse the impact of these cases further, annotator-familiarity was added as a metric. 

\subsection{Statistics}

Pilot tasks indicate that the compositionality and domain-specificity of NCCs vary between the two datasets, as non-compositional NCCs tended to occur more regularly in the Dúchas data (average ratio of 0.36) than in the UD-IDT (average ratio of 0.12), and domain-specificity and confidence scores of NCCs tended to be lower in Dúchas (average of 1.68 and 1.75) than in UD-IDT (average of 2.05 and 2.1). Examining the 270 NCCs annotated in the UD-IDT so far, compositionality scores average to 3.68, with 39 non-compositional (0-2) NCCs. Domain-specificity scores average to 1.8, with 224 NCCs annotated as non-domain-specific (1-2). Confidence scores average to 2.32, with 134 annotated as low-confidence (1-2). 

Based on this preliminary analysis, Dúchas data may provide a valuable resource for collecting NCs that do not overlap with terminology, however, non-standard orthography and low annotator familiarity may contribute towards more challenging cases, resulting in low annotator confidence.

\section{Conclusion and Future Work}

The findings of this ongoing work are limited by the scarcity of existing data, and the limitations of the annotators themselves. We also reduce the problem so as to minimise disagreements (i.e. only considering two-word NCs, removing NCs containing a definite article), which narrows the applications of this research. Nevertheless, this work represents a valuable contribution towards better understanding of NCs in Irish, and has potential applications to researchers in other languages who may be attempting a similar task.

We present our effort in building a corpus of Irish text annotated for noun compounds, with ratings of compositionality, domain-specificity, annotator familiarity and confidence. This work forms a critical basis towards developing Irish-specific NC resources, which will enable further evaluation of LLM capacity in a multilingual context. We present our annotation guidelines, insights gleaned from the annotation process, and some preliminary analysis. A collection of Irish NCs gathered during this effort will be released alongside the annotated corpus. Further annotation and analysis may shed more light on the question of how NCs present and the precise qualities that determine an NC in Irish.
%Discuss the eventual release of dúchas dataset
%MT experimental approach
%Applications of this research (insight into NCs in Irish, evaluate LLMs for these tricky Irish features)

\section{Acknowledgements}
“The Schools’ Collection, Volume 0621, Page 413” by Dúchas © National Folklore Collection, UCD is licensed under CC BY-NC 4.0. This work is funded in part by the Department of Tourism, Culture, Arts, Gaeltacht, Sport and Media.

\bibliography{custom}

\end{document}